%% file: main.tex
\documentclass[runningheads]{llncs}

\usepackage[mobile]{eccv}

\usepackage{eccvabbrv}

\usepackage{graphicx}
\usepackage{booktabs}

\usepackage[accsupp]{axessibility}  % Improves PDF readability for those with disabilities.

\usepackage{hyperref}
\begin{document}

\title{\textcolor{magenta}{Bi-}\textcolor{cyan}{TTA}: \textcolor{magenta}{Bi}directional \textcolor{cyan}{T}est-\textcolor{cyan}{T}ime \textcolor{cyan}{A}dapter for Remote Physiological Measurement} 
% Bi-TTA: Bidirectional Test-Time Adapter for Remote Physiological Measurement

\titlerunning{Bi-TTA}

\author{Haodong Li\inst{1} \and
Hao Lu\inst{1} \and
Ying-Cong Chen\inst{1,2}}
% \author{Under-review}

\authorrunning{H.~Li et al.}


\institute{Hong Kong University of Science and Technology (Guangzhou) \and
Hong Kong University of Science and Technology\\
\email{\{hli736, hlu585\}@connect.hkust-gz.edu.cn;\ yingcongchen@ust.hk}\\
\href{https://bi-tta.github.io}{\textcolor{magenta}{\fontfamily{cmtt}\selectfont{bi-tta.github.io}}}
}
% \institute{\url{https://bi-tta.github.io}}

\maketitle
\vspace{-0.8cm}
% \begin{figure}[t]
\begin{figure}
    \centering
    \includegraphics[width=0.31\linewidth]{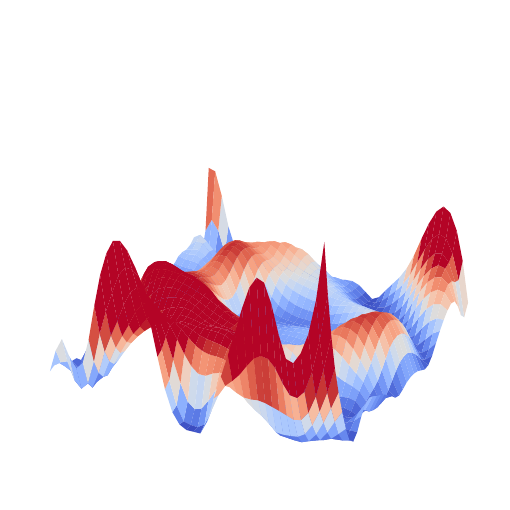} 
    \includegraphics[width=0.31\linewidth]{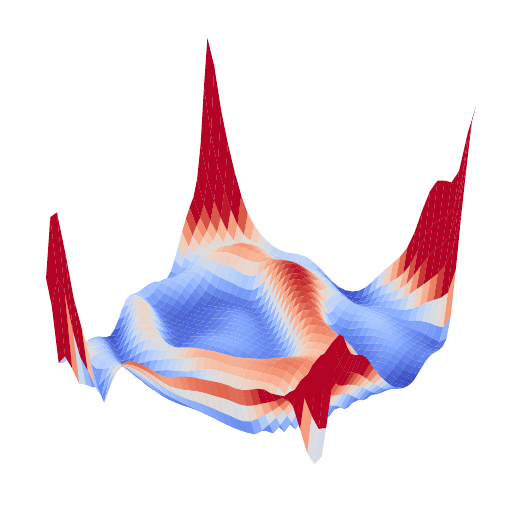} 
    \includegraphics[width=0.31\linewidth]{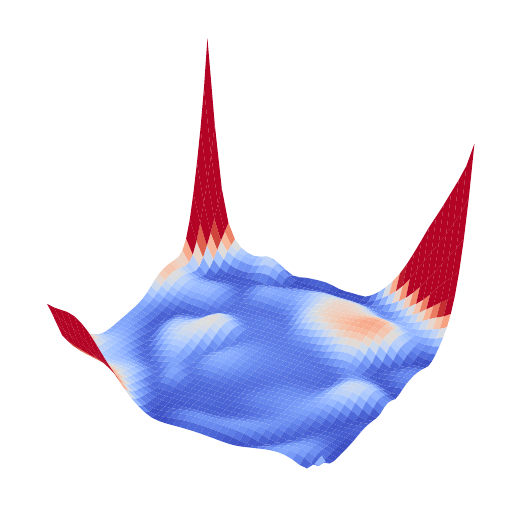} 
    \includegraphics[width=0.31\linewidth]{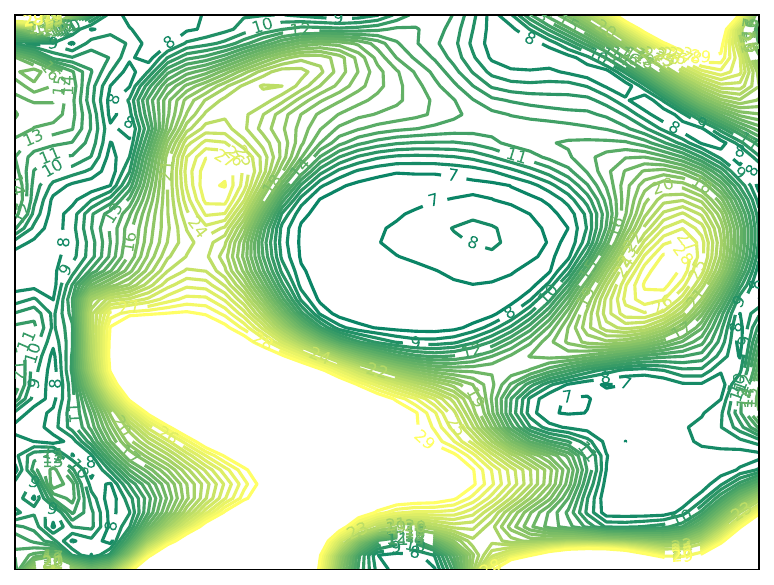}
    \includegraphics[width=0.31\linewidth]{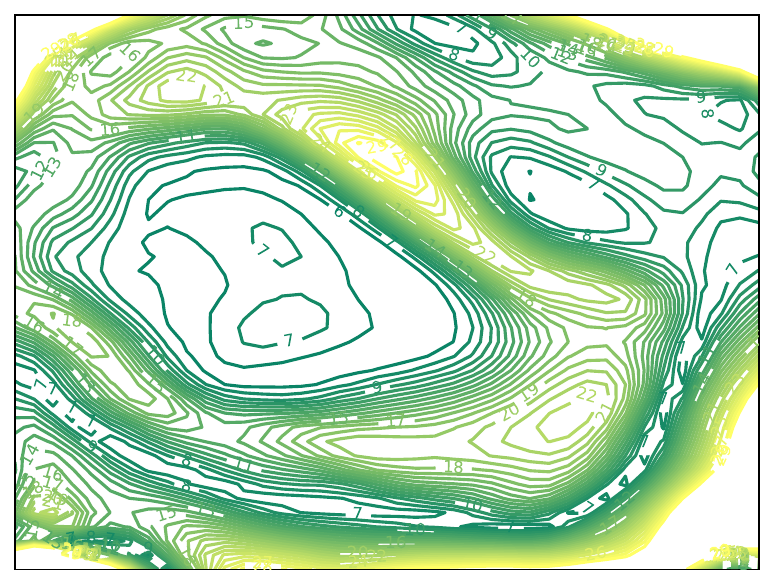}
    \includegraphics[width=0.31\linewidth]{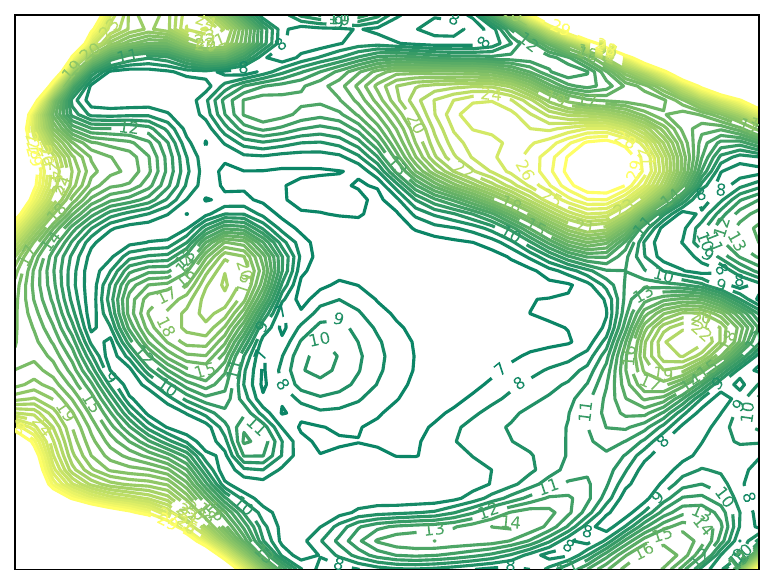}
    {
    \setlength{\fboxsep}{0pt}
    \setlength{\fboxrule}{0pt}
    \framebox[0.31\linewidth]{\small (a) Pre-trained}
    \framebox[0.31\linewidth]{\small (b) Priors-based TTA}
    \framebox[0.31\linewidth]{\small (c) Priors-based Bi-TTA}
    }
    % \vspace{0.3cm}
    \caption{Visualization of the mean absolute error (MAE) result on down-sampled VIPL dataset$\empty^{\ref{fn:vipl_20}}$ using \textbf{(a) the pre-trained rPPG model; (b) priors-based test-time adapted model; and (c) priors-based bidirectionally test-time adapted model.} These visualizations demonstrate that our proposed priors and the Bi-TTA framework significantly enhance the generalization performance (reflected by the flatness of MAE field~\cite{li2018visualizing}) beyond the pre-trained model$\empty^{\ref{fn:contour}}$.}
    \label{fig:contours}
\end{figure}
% \end{figure}
\vspace{-0.5cm}

\begin{abstract}
% \vspace{0.3cm}
% adapting the pre-trained model to target domain during inference, 
Remote photoplethysmography (rPPG) is gaining prominence for its non-invasive approach to monitoring physiological signals using only cameras. Despite its promise, the adaptability of rPPG models to new, unseen domains is hindered due to the environmental sensitivity of physiological signals. To address this issue, we pioneer the Test-Time Adaptation (TTA) in rPPG, enabling the adaptation of pre-trained models to the target domain during inference, sidestepping the need for annotations or source data due to privacy considerations. Particularly, utilizing only the user's face video stream as the accessible target domain data, the rPPG model is adjusted by tuning on each single instance it encounters. However, 1) TTA algorithms are designed predominantly for classification tasks, ill-suited in regression tasks such as rPPG due to inadequate supervision. 2) Tuning pre-trained models in a single-instance manner introduces variability and instability, posing challenges to effectively filtering domain-relevant from domain-irrelevant features while simultaneously preserving the learned information. To overcome these challenges, we present \textbf{Bi-TTA}, a novel expert knowledge-based \textbf{Bi}directional \textbf{T}est-\textbf{T}ime \textbf{A}dapter framework. Specifically, leveraging two expert-knowledge priors for providing self-supervision, our Bi-TTA primarily comprises two modules: a prospective adaptation (PA) module using sharpness-aware minimization to eliminate domain-irrelevant noise, enhancing the stability and efficacy during the adaptation process, and a retrospective stabilization (RS) module to dynamically reinforce crucial learned model parameters, averting performance degradation caused by overfitting or catastrophic forgetting. To this end, we established a large-scale benchmark for rPPG tasks under TTA protocol, promoting advancements in both the rPPG and TTA fields. The experimental results demonstrate the significant superiority of our approach over the state-of-the-art (SoTA).

\end{abstract}

\footnotetext[1]{This dataset is $1/20$ equally spaced down-sampled from the full VIPL dataset \cite{niu2018VIPL}, and contains 8349 samples in total.\label{fn:vipl_20}}

\footnotetext[2]{This visualization is generated following \cite{li2018visualizing}.\label{fn:contour}}
% where network weights were perturbed in a ratio ranging of $(-1, 1)$ in both  directions

\input{0305-GPT-Revised-HaodongLI}

\bibliographystyle{splncs04}
\bibliography{main}
\end{document}

%% file: 0305-GPT-Revised-HaodongLI.tex
\section{Introduction}
\label{sec:introduction}
The cardiac cycle refers to the process experienced by the cardiovascular system from the start of one heartbeat to the beginning of the next. Through the analysis of the cardiac cycle, important physiological indicators such as heart rate (HR), respiration rate (RF), and heart rate variability (HRV) can be derived, playing crucial roles in medical diagnostics, health surveillance, and forensic analysis~\cite{sun2016lstm,kessler2017pain,mcduff2022applications}.
% This information can effectively help us assess a person's health and emotional state, playing a crucial role in medical diagnosis, health monitoring, and forensic detection~\cite{sun2016lstm,kessler2017pain,mcduff2022applications}.

Traditional physiological measurement devices like electrocardiograms (ECG), heart rate bands, and finger-clips, despite their accuracy, are often expensive and uncomfortable to wear.
In contrast, rPPG offers a non-invasive and more convenient alternative by extracting blood volume pulses (BVP) from facial videos, analyzing the skin's light absorption variations to measure HR, HRV, and RF~\cite{yu2019remote,CVD2020,yu2021physformer}.
These monitors are especially important for tracking health status and sympathetic activity levels~\cite{sun2016lstm,kessler2017pain,mcduff2022applications}. Nowadays, physiological measurements based on rPPG, using only ordinary cameras, are increasingly becoming the focus of research in computer vision community~\cite{mcduff2021camera,2020PulseGAN,lu2021dual,bvpnet,niu2019robust,Qiu2019EVM, Du_2023_CVPR}.
However, the periodic color changes in the skin captured by remote cameras are quite subtle, and variables like natural scene illumination and head movements can significantly impair the reliability of physiological indicator measurements.
% very subtle, and factors such as natural scene lighting and subject head movements can greatly degrade the measurement relibility of physiological indicators.

Conventional rPPG methods~\cite{poh2010non,2014ICA,2011rPPGPCA, verkruysse2008remote_GREEN,de2013robust_CHROM,wang2017algorithmic_POS} perform well only in constrained environments. Recent deep learning (DL) methods~\cite{niu2019robust,revanur2021V4V,PURE2014,UBFC2017,xi2020BUAA} have demonstrate promising results,
however, the most of training data are collected in controlled laboratory environments with limited variations in illumination, camera parameters, etc. Therefore, these methods still struggle to generalize to unseen domains, necessitating model adaptation.

\begin{figure}[!t]
  \centering
   \includegraphics[width=0.75\linewidth]{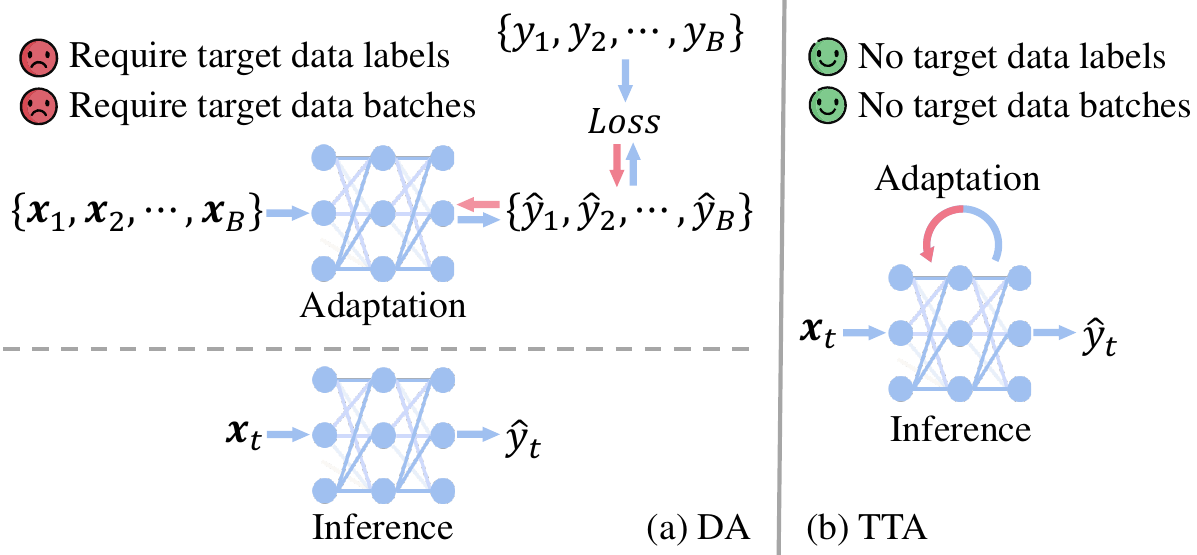}
   % \vspace{0.15cm}%final
   \caption{\textbf{Visualization of Domain Adaptation (DA) and Test-Time Adaptation (TTA) methodologies.} DA utilizes batch learning with labeled target data, while TTA dynamically refines the model during inference without relying on target data labels or distribution. Both DA and TTA obviate the requirement for source data.
    Note that domain generalization (DG) is excluded as it does not focus on \textit{specific} target domain adaptation.}
   \label{fig:Protocol}
\end{figure}

For privacy considerations, the adaptation of rPPG models in unseen scenarios is required to be performed without accessing the source data, and target data annotations or distributions (\textit{i.e.}, large batches of target data samples).
Addressing this, we firstly implement Test-Time Adaptation (TTA) in the field of rPPG.
TTA aims to fine-tune a pre-trained source model during inference time, as illustrated in Fig.~\ref{fig:Protocol}, without accessing the distribution and labeling of both source data and target data and naturally eliminate the need for intensive re-training~\cite{wang2021tent}. 
This innovation allows for the automatic adaptation of pre-trained rPPG models to unseen domains during inference, without the necessity for labelling or batches of target data. %, \textit{i.e.}, only the  is available.

However, applying TTA directly into rPPG faces two key limitations. 1) Predominantly designed for classification tasks, most TTA algorithms leverage entropy in normalization layers or pseudo labels. These strategies are not inherently suited for regression tasks like rPPG, leading to a lack of appropriate supervision. 
2) In real-world scenarios, the pre-trained rPPG model is tuned during inference the user's face video stream, on an instance-by-instance context.% of continual single samples. 
This variability and instability introduced by biased instances pose challenges to effectively filtering out domain-irrelevant noise and simultaneously preserving crucial learned information as the model acquiring new knowledge.
This nuanced balance is critical in preventing slow or ineffective adaptation
under single-instance learning condition, ensuring an efficient and consistent adaptation process.

To navigate these challenges, we proposed a novel \textbf{Bi}directional \textbf{T}est-\textbf{T}ime \textbf{A}dapter (\textbf{Bi-TTA}) framework, grounded on two expert knowledge-based self-supervised priors.
Firstly, two regularizations are proposed for offering effective supervision especially in rPPG tasks, addressing the aforementioned ill-suited problem of applying existing TTA algorithms into regression tasks. These two priors focus on temporal and spatial consistency respectively, which is derived from the characteristics of physiological signals.
Then, two strategies are introduced for stable and effective fine-tuning process. 
The prospective adaptation (PA) strategy is designed for filtering out the irrelevant information of each encountered instance, aiming that only target-domain-essential feature is truly contributing to model adaptation.
The retrospective stabilization (RS) strategy backtracks and consolidates essential model parameters dynamically, leveraging previous update gradients during the self-supervised tuning process, ensuring that the learned adaptation ability is preserved intact. % using the existing tuning gradient
We evaluate our method across various datasets, benchmarking against existing prevailing TTA algorithms, demonstrating the superiority of our Bi-TTA. Overall, our contributions can be summarized as follows:
\begin{itemize}
    \item {We introduce the TTA protocol to rPPG tasks, paving the way for a new paradigm in self-supervised adapting of rPPG models during inference.}
    \item {We introduce two novel self-supervised priors to regularize temporal and spatial consistency, providing appropriate supervision.}
    \item {We propose a novel Bidirectional Test-Time Adapter (Bi-TTA) framework grounded on the introduced priors, which promotes effective and efficient adaptation, with enhanced stability.}
    \item {We established a large-scale benchmark for rPPG tasks under TTA protocol. Extensive experiments demonstrate significantly improved adaptation capacity of our approach.}
\end{itemize}

\section{Related Works}
\label{sec:related_works}
\subsection{Remote Physiological Measurement}
Remote physiological measurement aims to quantify HR and HRV from BVP signals by interpreting skin chrominance fluctuations. Traditional techniques typically involve color space transformations or signal decomposition to obtain the BVP signal with high signal-to-noise ratio (SNR)~\cite{de2013robust_CHROM,wang2015exploiting,de2014improved,poh2010non}. With powerful modeling capacity of deep learning (DL), there's been a significant shift towards employing DL models for remote physiological estimation, yielding impressive results.~\cite{Chen_2018_ECCV,hsu2017deep,synrhythm,niu2019robust,vspetlik2018visual,wang2019vision,yu2019remote}. % noteworthy
Many of these methods extract BVP signals from face videos typically by aggregating spatial and temporal information~\cite{vspetlik2018visual,yu2019remote,CVD2020,yu2021physformer}. Besides, many works utilize hand-designed 2D feature maps extracted from face videos to alleviate computational demands ~\cite{niu2019robust,song2020heart,hsu2017deep,reiss2019deep,lu2021dual}. Recently, researchers have turned to exploring how to leverage task-specific knowledge and how to improve the generalization performance of the model into unseen scenarios.

% \vspace{0.1in} 
\noindent \textbf{Task-specific priors.} rPPG is distinguished by its unique biological and physical mechanisms, which can be exploited to enhance the physiological signal extraction ability of the model~\cite{wang2022self,gideon2021way,sun2022contrast,liu2023rppg}. 
These task-specific priors can be summed up as 1) spatial consistency: BVP signals extracted from different face areas should exhibit similar; 2) temporal consistency: BVP signals typically exhibit smooth and gradual changes over short time period; 3) heart rate range: The heart rate typically ranges between 40 and 250 beats per minute (bpm). Different from existing works, we provide a more granular solution leveraging the multi-scale latent features, rather than purely focusing on the prediction output of BVP signals.

% \vspace{0.1in} 
\noindent \textbf{The generalization of the model.} With the domain shifts (\textit{e.g.}, lighting, head motion, environment, etc.), most methods tend to have significant performance degradation. To improve the generalization performance on unseen domains, NEST-rPPG established a large-scale DG evaluation protocol~\cite{Lu_2023_CVPR}. These DG methods, requiring the access of all source data, do not focus on the adaptation of a specific target domain, leading to sub-optimal performance~\cite{sun2023resolve}. Besides, some work utilizes the labels and distribution of the target domain data to improve the model's adaptation performance~\cite{lee2020meta,liu2021metaphys,du2023dual}, which is impractical due to accessing the unseen domain in advance. Differently from the above rPPG methods, we address this domain gap by using the TTA protocol, facilitating the adaptation to specific target domains during inference, based solely on unlabeled instances from the target domain. % and conduct the target domain adaptation during inference.

\subsection{Test-Time Adaptation}
Test-time Adaptation (TTA), an emerging protocol, aims to adjust a pre-trained source model during inference by learning from the unlabeled target data, without accessing the distributions and labels of both source and target data, thereby naturally eliminate the need of intensive re-training~\cite{wang2021tent, wang2022generalizing, liang2023comprehensive}. Current TTA methods typically require aggregating a large number of samples from the target data to form a batch, and perform adaptation through 1) normalization calibration~\cite{klingner2022continual, bahmani2022semantic, zou2022learning, khurana2021sita, niu2022efficient}, 2) pseudo-labeling~\cite{gandelsman2022testtime, d2019learning}, 3) input adaptation~\cite{zhao2022test, gao2023back}, 4) dynamic inference that learns the network weights from a set of model parameters~\cite{zhang2023adaptive}, 5) meta-learning~\cite{pmlr-v70-finn17a}, etc. These approaches can be considered as different forms of 
source-free domain adaptation (SFDA)~\cite{liang2020we} and are infeasible for regression tasks like rPPG where data arrives continuously and in a sequential manner. 
Some instance-level TTA approaches may either rely on specialized network architectures~\cite{klingner2022continual, bahmani2022semantic, wang2021tent, zou2022learning} or incorporate additional auxiliary tasks applied during both pre-training and adaptation~\cite{d2020one, sun2020test}.
By imposing specific non-general requirement to the pre-training stage, these methods narrow application scope of TTA.
Moreover, some works even introduce a memory mechanism that turns instances into batches~\cite{khurana2021sita}, lacking an in-time adaptation to each inference with increased computational cost.
In contrast with above mentioned TTA methods, our Bi-TTA adapts the model during inference under each encountered instance from target domain without specific pre-training or network architecture requirements, offering a flexible, effective, and efficient solution for rPPG model adaptation.

\section{Method}
\label{sec:method}
\subsection{Overview}
In this paper, we propose a novel expert knowledge-based Bi-directional Test-Time Adapter (Bi-TTA) to address the performance degradation in unseen target domains. In the following sections, we will firstly formulate the problem in Sec.~\ref{sec:problem}. Secondly, Sec.~\ref{sec:prior} introduces the proposed self-supervised priors based on expert knowledge in the realm of rPPG. The bidirectional (prospective and retrospective) adaptation strategy of our Bi-TTA will be specified in Sec.~\ref{sec:bitta}.

\subsection{Preliminaries}
\label{sec:problem}
The source dataset $D_\text{S} \triangleq \cup_{i=1}^{n_\text{S}}\left\{\left(\boldsymbol{x}_i, {y}_i\right)\right\}$ is assumed to be drawn independently and identically distributed (i.i.d.) from distribution $X_\text{S}$. similarly, the target dataset $D_\text{T}$ is assumed to be sampled from a different distribution $X_\text{T}$, reflecting the domain shift. %  \triangleq \cup_{i=1}^{n_{\text{T}}}\left\{\boldsymbol{x}_i\right\}
With a model parameterized by $\boldsymbol{w} \in \mathcal{W}$ and considering the supervised loss function $L: \mathcal{W} \times \mathcal{X} \times \mathcal{Y} \rightarrow \mathbb{R}_{+}$ for evaluation, we define the observation loss over dataset $D_\text{}$ as an approximation of the population loss over distribution $X_\text{}$:
\begin{equation}
\begin{aligned}
\mathcal{L}_{D_\text{}}(\boldsymbol{w}) = \frac{1}{n} \sum_{i=1}^{n_\text{}} L\left(\boldsymbol{w}, \boldsymbol{x}_i, {y}_i\right)\approx  \mathbb{E}_{(\boldsymbol{x}, {y}) \sim {X}_\text{} }\left[L(\boldsymbol{w}, \boldsymbol{x}, {y})\right]\text{.}
\end{aligned}
\label{eq:01}
\end{equation}
The objective of TTA, including our Bi-TTA, is to minimize the target observation loss $ \mathcal{L}_{D_\text{T}}(\boldsymbol{w}_t) $ after fine-tuning the initial model weights from $\boldsymbol{w}_{0}$ to $\boldsymbol{w}_{t}$ after $t$ observed samples.

\begin{figure}[!t]
\centering
\subcaptionbox{STMap Construction \label{fig:stmap}}{
\includegraphics[width = 0.28\linewidth]{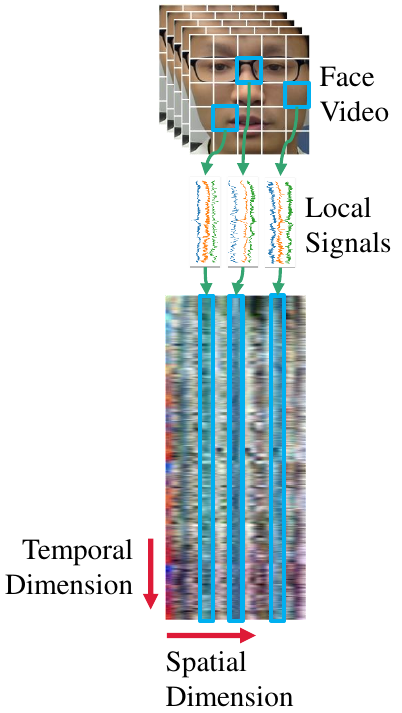}
}
\hfill
\subcaptionbox{\scalebox{0.9}{Temporal Consistency Loss} \label{fig:tcl}}{
\includegraphics[width = 0.28\linewidth]{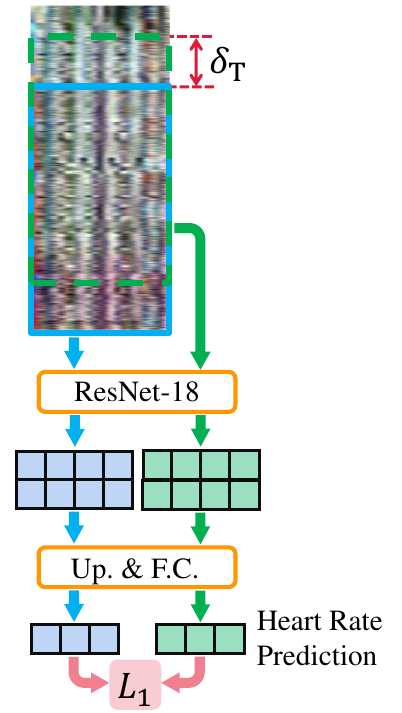}}
\hfill
\subcaptionbox{Spatial Consistency Loss \label{fig:scl}}{
\includegraphics[width = 0.38\linewidth]{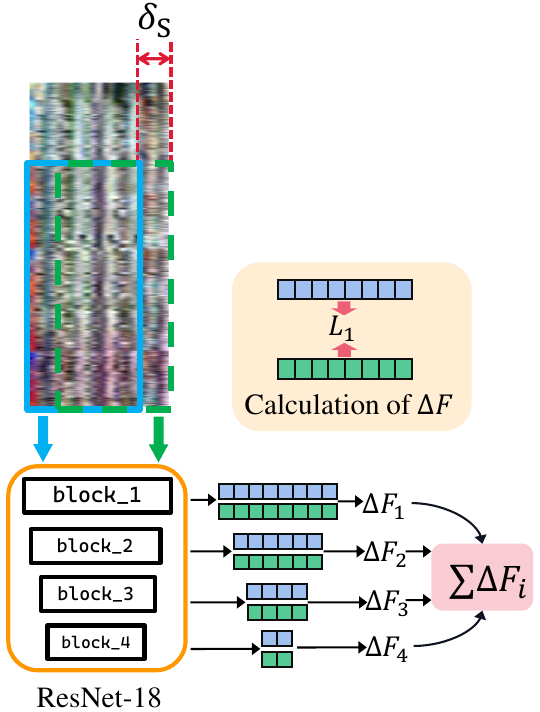}}
\caption{\textbf{Illustration of STMap construction and the implementation of our proposed expert knowledge-based priors.} (a) The process of generating STMap, encompassing face alignment$\empty^{\ref{fn:face}}$ and cropping, local signal extraction, and the subsequent integration. (b) The calculation process of TCL, aimed at minimizing significant prediction discrepancies between original and temporally shifted HR predictions. (c) The calculation process the SCL, focused on penalizing pronounced disparities across different facial regions. Note that the boxes colored in \textcolor{magenta}{pink} represent the loss outcomes.}
\label{fig:priors}
\end{figure}

\subsection{Expert Knowledge-based Priors}
\label{sec:prior}
In the context of Bi-TTA, the unavailability of target data labels compels the use of self-supervised loss functions for providing the tuning gradient. Inspired by~\cite{wang2022self,gideon2021way,sun2022contrast,liu2023rppg}, our Bi-TTA capitalizes on the characteristics of rPPG signals to craft these self-supervised loss functions, concentrating on the principles of spatial and temporal consistency. The introduced loss functions are termed as $L_{\text{s}}\left(\boldsymbol{w}_{t-1}, \boldsymbol{x}_t\right)$ and $L_{\text{t}}\left(\boldsymbol{w}_{t-1}, \boldsymbol{x}_t\right)$, focusing on spatial and temporal consistency respectively.
These functions play a foundational role in the self-supervised tuning process, providing appropriate supervision based on the expert knowledge, despite the lack of direct label guidance.

% \vspace{0.1in} 
\noindent \textbf{Spatial-temporal feature map (STMap).} In line with~\cite{niu2019robust,song2020heart,hsu2017deep,reiss2019deep,lu2021dual}, 2D STMap is firstly constructed to encapsulate the face video, as specified in Fig.~\ref{fig:stmap}. This STMap integrates both temporal and spatial dimensions: the temporal dimension represents the time sequence and the spatial dimension corresponds to the patches of the cropped face video frames.
The current sample $ \boldsymbol{x}_t \in \mathbb{R}^{W \times H} $ is referred to as a sliding window on the STMap, where $W$ represents the sequential length of the sliding window, and $H$ indicates the size of spatial dimension in the STMap.

% \vspace{0.1in} 
\noindent \textbf{Temporal consistency loss (TCL).} 
BVP signals exhibit smooth and gradual fluctuations over short time spans. This fundamental attribute, known as the temporal consistency, is essential for accurate physiological signal analysis and the key to  guarantee the predictability of physiological signal changes. To preserving this critical characteristic and derive a tuning gradient accordingly, the TCL is meticulously designed to enforce the model's prediction deviations of BVP signals when subjected to minor time perturbations under an acceptable tolerance. To compute the TCL, a random minor temporal perturbation $\delta_{\text{T}}$ is introduced by shifting the original sample $\boldsymbol{x}_t$ along the STMap's temporal dimension. The shifted sample $\boldsymbol{x}_{t-\delta_{\text{T}}}$, in parallel with the origin sample, are then fed into the rPPG network $\boldsymbol{w}_{t-1}$ for predicting two sets of BVP signals.
Then, for penalizing the model from large discrepancies in the HR estimation results between the original and shifted samples, $L_1$ regularization is applied on the original and shifted HR prediction results, incorporating with a tolerance constant $ \xi_{\text{T}} $.
The specific calculation process of TCL is formulated in Eq.~\ref{eq:02}, as illustrated in Fig.~\ref{fig:tcl}.

\begin{equation}
\begin{aligned}
    \Delta_{t,t-\delta_{\text{T}}}^{\text{HR}} &= \left\| \boldsymbol{w}_{t-1}^{\text{HR}} (\boldsymbol{x}_t) - \boldsymbol{w}_{t-1}^{\text{HR}} (\boldsymbol{x}_{t - \delta_{\text{T}}}) \right\|_1 \in \mathbb{R}^{W\times 1}\\
    L_{\text{t}}\left(\boldsymbol{w}_{t-1}, \boldsymbol{x}_t\right) &=  \sum_{i}^{W} \max \left(0, \Delta_{t,t-\delta_{\text{T}}}^{\text{HR}(i)} - \xi_{\text{T}} \right).
\end{aligned}
\label{eq:02}
\end{equation}
\footnotetext[3]{\href{https://github.com/1adrianb/face-alignment}{https://github.com/1adrianb/face-alignment}.\label{fn:face}}

% \vspace{0.1in} 
\noindent \textbf{Spatial consistency loss (SCL).} 
Heart rate variations induce chromatic changes across the face, which are periodic and consistent across different facial regions. This phenomenon is referred to as spatial consistency, alongside temporal consistency, are crucial attributes in the rPPG field.
Inspired by this phenomenon, SCL is crafted to promote the spatial consistency, obtaining correct adaptation gradient by encouraging the BVP signals from different facial areas to exhibit similarities.
In particular, the SCL is calculated covering a series of latent feature maps,
corresponding to various depths, (\textit{i.e.}, the number of residual blocks) in the adopted ResNet-18~\cite{he2016deep} architecture.
Given the current sample $ \boldsymbol{x}_t \in \mathbb{R}^{W \times H} $, four latent feature maps $\left\{ F_i\in \mathbb{R}^{W_i\times H_i}\ |\ i \in [1,4]\right\} $ could be extracted from the rPPG network $\boldsymbol{w}_{t-1}$. As illustrated in Fig.~\ref{fig:scl}, the calculation process of SCL is specify as:
\begin{equation}
L_{\text{s}} =\sum_{i}^n  \Delta F_i = \sum_{i}^n \sum_{j}^{W_i - \delta_{\text{S}}} || F_{i, j} - F_{i, j + \delta_{\text{S}}} ||_1, \\
\label{eq:03}
\end{equation}
where $ F_{i,j} = F_i[:,j]\in \mathbb{R}^{W_i} $, $n=4$ represents the number of residual blocks, and $\delta_S$ is the incremental spatial shift. Leveraging the fine-grained multi-level strategy, SCL effectively nurtures a comprehensive regularization of spatial features, ensuring that appropriate supervision is available for the adaptation during inference without annotations.

After combining $L_{\text{t}}$ and $L_{\text{s}}$, without direct target label reliance, the self-supervised loss is formulated by weighting and summing the spatial and temporal consistency priors as follows:
\begin{equation}
L_{\text{p}} = \lambda_{\text{s}}L_{\text{s}} + \lambda_{\text{t}}L_{\text{t}}\text{.}
    \label{eq:06}
\end{equation}
\subsection{Bidirectional Test-Time Adaptation}
\label{sec:bitta}
\definecolor{kellygreen}{rgb}{0.3, 0.73, 0.09}
\begin{figure}[!t]
  \centering
   \includegraphics[width=0.5\linewidth]{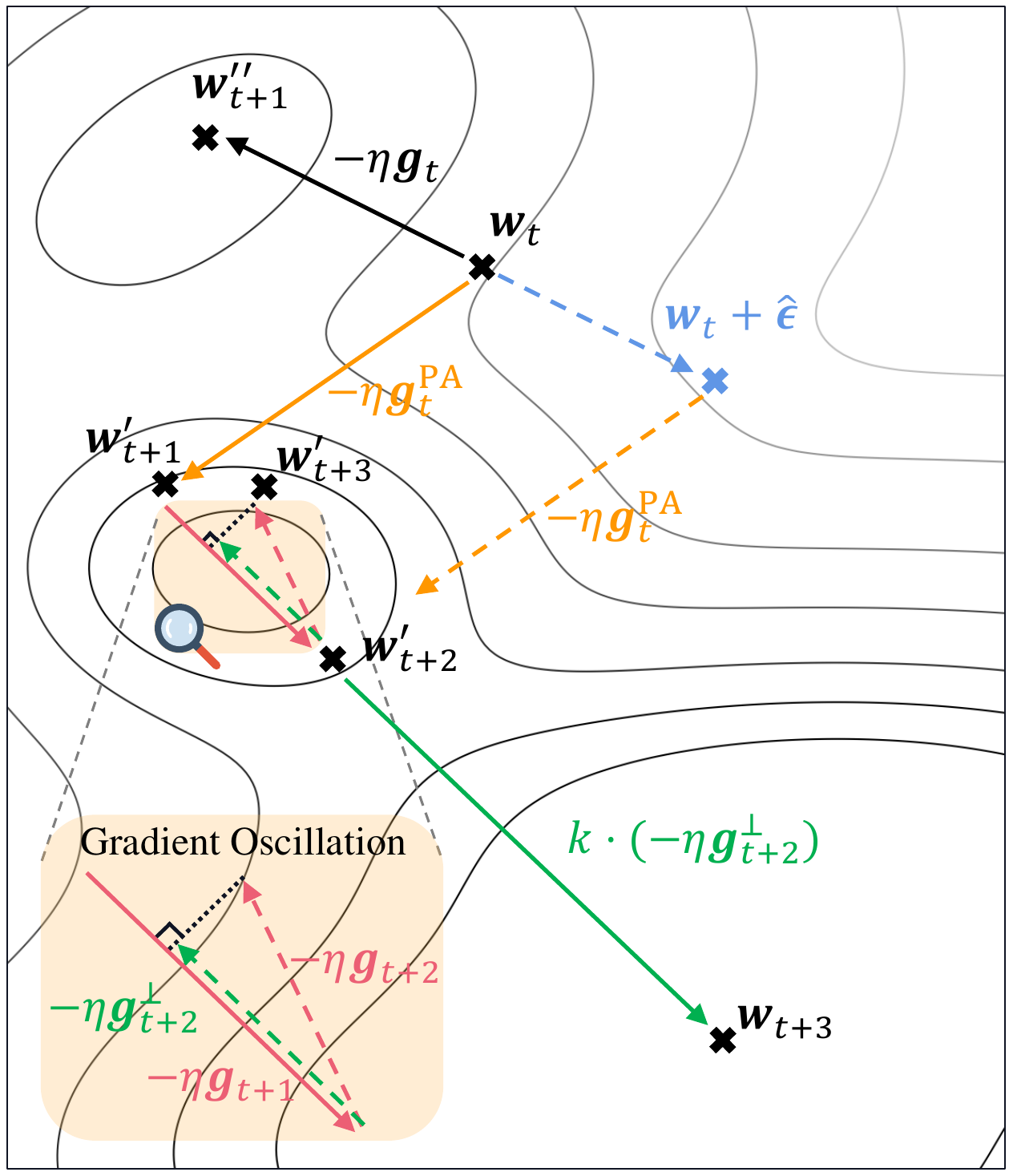}
   % \vspace{0.15cm}%final
   \caption{
\textbf{Illustration of the proposed Bidirectional Test-Time Adapter (Bi-TTA).} Black arrows $\rightarrow$ indicate the adaptation process purely with the proposed two priors, \textit{i.e.}, TCL and SCL. Orange arrows \textcolor{orange}{$\rightarrow$} denote that the PA module adjusts model parameter using the gradient of representative neighborhood with a radius $\rho$. The green ones \textcolor{kellygreen}{$\rightarrow$} show that the RS is activated when there is an oscillation, which is a sign of performance degradation, for maintaining the essential learned adaptation ability with former tuning gradients.
The gradient and learning rate are formulated as $\boldsymbol{g}$ and $\eta$ respectively.
}
   \label{fig:bitta}
\end{figure}

As introduced above, based on the proposed priors, the Bi-TTA framework aims to ensure effective and robust model adaptation under the variability and instability of single-instance learning conditions, minimizing the impact of domain-irrelevant noise. 
Simultaneously adapting the model with new target samples, our Bi-TTA also focuses on preserving the acquired adaptation ability dynamically, preventing the performance degradation caused by overfitting or catastrophic forgetting, achieving a stable maintenance of the adaptation performance in the target domain.

% \vspace{0.1in} 
\noindent \textbf{Prospective adaptation (PA).}
During the process of model adaptation, especially under single-instance learning conditions, the noisy irrelevant information in biased target samples can easily interfere the model's tuning process, leading to a declined performance. Therefore, inspired by~\cite{mi2022make, foret2021sharpnessaware}, rather than seeking out model parameter $\boldsymbol{w}_{t}$ that simply have low adaptation loss $L_{\text{p}}\left(\boldsymbol{w}_t\right)$, we seek out model parameter whose entire neighborhoods to have uniformly low adaptation loss. 
\begin{equation}
      \label{eq:pa1}
      \begin{aligned}
      {L}_{\text{p}}'(\boldsymbol{w}) = \max _{\|\boldsymbol{\epsilon}\|_2  \leqslant  \rho} L_{\text{p}}(\boldsymbol{w}+\boldsymbol{\epsilon}),
      \end{aligned}
  \end{equation}
where the $\rho$ is the disturbance range. This strategy, known as sharpness-aware minimization, has been proved to be effective in various areas~\cite{wu2020adversarial,jia2021scaling,pham2021meta,brock2021high}.
For computing efficiency and differentiability, we approximate the $\boldsymbol{\epsilon}^*= \underset{\|\boldsymbol{\epsilon}\|_2 \leqslant \rho}{\arg\max } \ L_{\text{p}}(\boldsymbol{w}+\boldsymbol{\epsilon})$.
Following $1$-order Taylor series, we get:
% $f(x) \approx f(a) + f'(a)(x-a)$ w.r.t. $x$ around $a$.
% $ = \underset{\|\boldsymbol{\epsilon}\|_2 \leqslant \rho}{\arg\max } \ L_{\text{p}}(\boldsymbol{w}+\boldsymbol{\epsilon})   $
\begin{equation}
\begin{aligned}
    \boldsymbol{\epsilon}^* &\approx \underset{\|\boldsymbol \epsilon\|_2 \leqslant \rho}{\arg \max } L_{\text{p}}(\boldsymbol{w})+\boldsymbol{\epsilon}^\text{T} \nabla_{\boldsymbol{w}} L_{\text{p}}(\boldsymbol{w})\\&=\underset{\|\boldsymbol{\epsilon}\|_2 \leqslant \rho}{\arg \max } \boldsymbol{\epsilon}^\text{T} \nabla_{\boldsymbol{w}} L_{\text{p}}(\boldsymbol{w}) =\underset{\|\boldsymbol{\epsilon}\|_2 \leqslant \rho}{\arg \max } \boldsymbol{\epsilon}^\text{T} \boldsymbol{g}=\hat{\boldsymbol{\epsilon}},
\end{aligned}
 \end{equation}
, w.r.t. $\rho$ around $0$.
Then employing the Cauchy-Schwarz inequality from Dual Norm theory: 
% (here is a brief \href{https://inst.eecs.berkeley.edu/~ee127/sp21/livebook/def_dual_norm.html}{introduction}). 
% Our aim is to maximize $ \boldsymbol{\epsilon}^{\mathrm{T}}\nabla_{\boldsymbol{w}} L_{\text{p}}(\boldsymbol{w}) $ and meanwhile assuring $ \|\boldsymbol{\epsilon}\|_2 \leqslant \rho $. 
$\boldsymbol{\epsilon}^{\mathrm{T}} \boldsymbol{g} \leqslant\|\boldsymbol{\epsilon}\|_2\left\|\boldsymbol{g} \right\|_2$. $\boldsymbol{\epsilon}$ should be col-linear with $ \boldsymbol{g} $ to maximize $ \boldsymbol{\epsilon}^\text{T} \boldsymbol{g} $. Considering $ \|\boldsymbol{\epsilon}\|_2 \leqslant \rho $, we can get: 
\begin{equation}
\hat{\boldsymbol{\epsilon}}= \rho \frac{\operatorname{sign}(\boldsymbol{g})\boldsymbol{g}}{||\boldsymbol{g}||_2}=\rho \frac{\operatorname{sign}\left(\nabla_{\boldsymbol{w}} L_{\text{p}}(\boldsymbol{w})\right)\nabla_{\boldsymbol{w}} L_{\mathrm{p}}(\boldsymbol{w})}{\left\|\nabla_{\boldsymbol{w}} L_{\mathrm{p}}(\boldsymbol{w})\right\|_2}.
\end{equation}
% leveraging the first-order Taylor Series: $\hat{\boldsymbol{\epsilon}}=\rho \frac{\operatorname{sign}\left(\nabla_{\boldsymbol{w}} L_{\text{p}}(\boldsymbol{w})\right)\nabla_{\boldsymbol{w}} L_{\mathrm{p}}(\boldsymbol{w})}{\left\|\nabla_{\boldsymbol{w}} L_{\mathrm{p}}(\boldsymbol{w})\right\|_2}$. 
Consequently, we can derive the gradient approximation under Eq.~\ref{eq:pa1}:
\begin{equation}
      \label{eq:pa_2}
      \begin{aligned}
      \boldsymbol{g}^{\text{PA}}_t&=\nabla_{\boldsymbol{w}} {L}_{\text{p}}'(\boldsymbol{w})
\approx \nabla_{\boldsymbol{w}} L_{\mathcal{S}}(\boldsymbol{w}+\hat{\boldsymbol{\epsilon}})\\&=\left.\frac{d(\boldsymbol{w}+\hat{\boldsymbol{\epsilon}})}{d \boldsymbol{w}} \nabla_{\boldsymbol{w}} L_{\mathcal{S}}(\boldsymbol{w})\right|_{\boldsymbol{w}+ \hat{\boldsymbol{\epsilon}}} \approx \left.\nabla_w L_{\mathcal{S}}(\boldsymbol{w})\right|_{\boldsymbol{w}+\hat{\boldsymbol{\epsilon}}}
      \end{aligned}
  \end{equation}
As illustrated in Fig.~\ref{fig:bitta}, the gradient $\boldsymbol{g}^{\text{PA}}_t$ is adopted to replace the original gradient merely derived from two self-supervised priors. 
This strategy boosts the model's robustness to the undesired variability from instance-level conditions, 
safeguarding the tuning process against the influence from target-domain-irrelevant information.
Consequently, it secures both the efficacy and stability of the adaptation process.

% \vspace{0.1in} 
\noindent \textbf{Retrospective stabilization (RS).}
The PA module, by considering the model's entire neighborhood as the optimization objective, effectively enhances the model's robustness against irrelevant disturbances in target samples. 
However, after the model has processed a certain number of samples, it may still capture the detrimental noise feature instead of latent, universally domain-essential patterns, leading to performance degradation, as evidenced in Fig.~\ref{fig:abl}.
Therefore, to ensure reliable physiological signal predictions throughout the entire adaptation process, we propose the retrospective stabilization (RS) module. RS initially introduces the concept of \textit{trend gradient}, termed as $\boldsymbol{g}^{*}$. Starting with $\boldsymbol{g}_0^{*}=\mathbf{0}$, the trend gradient after processing $t$ samples is updated via: $\boldsymbol{g}^{*}_t =  \frac{\boldsymbol{g}^{*}_{t-1} + (1/||L_{\text{p}}||_1)\boldsymbol{g}_t^{\text{PA}}}{1 + 1/||L_{\text{p}}||_1}$. For each $\boldsymbol{g}^{\text{PA}}_t$, its orthogonal projection on $\boldsymbol{g}^{*}_t$ is computed, formulated as $ \boldsymbol{g}^{\perp}_t = \frac{\boldsymbol{g}^{\text{PA}}_t \cdot \boldsymbol{g}^{*}_t }{\left\| \boldsymbol{g}^{\text{}*}_t \right\|^2}  \boldsymbol{g}^{*}_t  $. 
As illustrated in Fig.~\ref{fig:bitta}, RS is activated if this projection aligns in the opposite of the trend gradient, which signifies an gradient oscillation, indicative of a potential fall of the generalization performance. Specifically, upon detection of oscillation, the model's optimization gradient will be the product of $ \boldsymbol{g}^{\perp}_t $ and the backtracking degree $k$. The orthogonal part of $ \boldsymbol{g}^{\text{}}_t $ on $\boldsymbol{g}^{*\text{}}_t$ is ignored.
\begin{equation}
    \label{eq:-1}
     \boldsymbol{g}_t^{\text{RS}}  = \displaystyle\left\{
\begin{aligned}
     &  k \cdot \boldsymbol{g}^{\text{}\perp}_t,\ \cos(\boldsymbol{g}^{*}_t, \boldsymbol{g}^{\text{PA}}_t) <0\\
     & (1-\lambda_t^{\text{RS}} )\boldsymbol{g}^{\text{PA}}_t + \lambda_t^{\text{RS}} \boldsymbol{g}^{\text{}*}_t ,\ \cos(\boldsymbol{g}^{*}_t, \boldsymbol{g}^{\text{PA}}_t) \geqslant 0 
\end{aligned}
\right.
\end{equation}
where $\lambda_t^{\text{RS}}= -1 + 2\cdot\operatorname{Sigmoid}(-\frac{4t}{\Omega}) \in \mathbb{R}_+^{(0,1)}$,  is the anneal scaling factor, $\Omega$ denotes the amount of target samples needed to ensure the fidelity of trend gradient, \textit{i.e.}, $\lambda_t^{\text{RS}}$ can be approximated as $1.0$ when $t\geqslant \Omega$. 
The obtained $\boldsymbol{g}_t^{\text{RS}}$ represents the gradient that ultimately contributes to fine-tune the model for target domain adaptation.
% The obtained $\boldsymbol{g}_t^{\text{RS}}$ is the gradient finally contributes to the model tuning for target domain adaptation.

\section{Experiments}
\label{sec:exp}

\begin{table*}[!t]
\centering
\scriptsize
\caption{
\textbf{HR estimation results.} Typically, the model selected for adaptation on each dataset is initially pre-trained on the remaining four datasets utilizing the NEST~\cite{Lu_2023_CVPR} framework. Ours w/o P.R. denotes that only the expert knowledge-based priors is adopted, without the bidirectional adaptation strategy. The best results are highlighted in \textbf{bold}, and the second-best results are \underline{underlined}.}
\setlength{\tabcolsep}{0.040cm}
\begin{tabular}{cc|c|c|c|c|c|c|c|c|c|c|c|c|c|c}
\toprule 
&\multicolumn{3}{c}{VIPL~\cite{niu2018VIPL}}     & \multicolumn{3}{c}{V4V~\cite{revanur2021V4V}}  & \multicolumn{3}{c}{PURE~\cite{PURE2014}}  & \multicolumn{3}{c}{UBFC~\cite{UBFC2017}}   & \multicolumn{3}{c}{BUAA~\cite{xi2020BUAA}}  \\
% & \multicolumn{3}{c}{UBFC~\cite{UBFC2017}}    & \multicolumn{3}{c}{PURE~\cite{PURE2014}}    & \multicolumn{3}{c}{BUAA~\cite{xi2020BUAA}}    & \multicolumn{3}{c}{VIPL~\cite{niu2018VIPL}}     & \multicolumn{3}{c}{V4V~\cite{revanur2021V4V}}      \\
\cmidrule(lr){2-4} \cmidrule(lr){5-7} 
\cmidrule(lr){8-10} \cmidrule(lr){11-13}  \cmidrule(lr){14-16} 
Method & M$\downarrow$  & R↓ & r↑    & M↓  & R↓ & r↑    & M↓  & R↓ & r↑    & M↓  & R↓  & r↑    & M↓  & R↓  & r↑    \\
\midrule
\multicolumn{16}{c}{\textit{Methods based on \textbf{DG} protocol}} \\
\midrule
Coral~\cite{coral}  & 8.68& 11.91& 0.53& 10.32& 14.42& 0.32& 7.59& 10.87& 0.72& 5.89& 8.04& 0.76& 3.64& 5.74& 0.8
\\
VREx~\cite{VREx}      & 8.37& 11.62& 0.54& 9.82& 14.16& 0.37& 7.24& 10.14& 0.78& 5.59& 7.68& 0.81& 3.27& 5.01& 0.86
\\
NCDG~\cite{NC2022PAMI}      & 8.47& 11.81& 0.52& 10.14& 14.46& 0.34&7.32& 10.35& 0.77& 5.31& 7.56& 0.82& 3.12& 5.16& 0.85
\\
NEST \cite{Lu_2023_CVPR}   & 7.86& 11.15& 0.58& 9.27& 13.79& 0.41& 6.71& 9.59& 0.81& 4.67& 6.79& 0.86& 2.88& 4.69& 0.89
\\
\midrule  \multicolumn{16}{c}{\textit{Methods based on \textbf{TTA} protocol}} \\
\midrule
Tent \cite{wang2021tent} & 8.09&11.32&0.55& 9.98&11.73& 0.47& 6.86& 9.68& 0.84&4.57&9.74& 0.87& 2.37& 3.59&0.92
\\
SAR \cite{niu2023towards} &8.13&11.45&0.53& 9.68& 11.46& 0.47&6.33& 9.3& 0.86&4.56& 9.64& 0.87&2.07& 2.93&0.95
\\
TTT++ \cite{liu2021ttt++} &7.73& 11.08& 0.56& 9.51& 11.31& 0.54&6.23&9.14& 0.88& 4.36& 6.79& 0.87&2.05&2.82&0.92
\\
EATA \cite{niu2022efficient} &7.69& 11.03& 0.57&9.36&11.12& 0.51& 6.13& 9.23& 0.86& 4.25& 6.86& 0.88& 1.89& 2.64& 0.94
\\
SHOT \cite{liang2021source} & 7.75& 11.07& 0.56&9.55&11.32&0.51&5.81&8.86& 0.87&4.05&6.45&0.87&1.87& 2.47& 0.96
\\
AdaODM \cite{zhang2023adaptive} & 7.81& 11.21& 0.55&9.1& 10.93& 0.55&5.51& 8.29& 0.88&4.01& 6.59& 0.87&1.91& 2.6& 0.95
\\
\midrule
ConPhys$\empty^{\ref{fn:conphys}}$ \cite{sun2022contrast} & 7.43& 10.7& 0.61& 9.3& 10.96& 0.52&6.09& 8.83& 0.88& 3.92& 6.61& 0.886& 1.75& 2.19& 0.97
\\
\textbf{Ours w/o P.R.} &7.31& 10.64&0.62& \underline{8.97}& \underline{10.82}& \underline{0.55} & 5.56& 8.67& 0.88& 3.64& 6.63& 0.885& 1.68& 2.04&0.99
\\
\midrule
\textbf{Ours w/o RS} &{\underline{7.15}} & {\underline{10.83}} &{\underline{0.63}}& \textbf{{8.93}}& \textbf{{10.79}}& \textbf{{0.55}}& \underline{5.24}& \underline{8.32}& \underline{0.889}&3.59&6.61&0.885& 1.55& 1.94& 0.99
\\
\textbf{Ours w/o PA} &7.24&10.78& 0.62& 9.04& 10.97& 0.54& 5.39&8.45& 0.88& \underline{3.54}& \underline{6.55} &\underline{0.888}& \underline{1.51}& \underline{1.98}&\underline{0.99}
\\
\textbf{Ours} & \textbf{{7.09}}& \textbf{{10.72}}&\textbf{{0.63}}&9.1&11.02&0.54& \textbf{{5.02}} &\textbf{{8.11}}& \textbf{{0.890}}&\textbf{3.53}& \textbf{6.24}&\textbf{0.891}& \textbf{{1.49}}&\textbf{{1.97}} &\textbf{{0.99}}
\\
\bottomrule 
\end{tabular}
\label{tab:exp}
\end{table*}
\footnotetext[4]{The proposed self-supervised loss in ConPhys~\cite{sun2022contrast} is applied for a comparative analysis of our proposed priors.\label{fn:conphys}}

\subsection{Datasets}
\label{sec:dataset}
We elaborately select five rPPG datasets, encompassing diverse races, levels of head motion, camera conditions, and ambient lighting, to establish a comprehensive large-scale benchmark of rPPG task under TTA protocol.

% \vspace{0.1in} 
\noindent \textbf{VIPL-HR}~\cite{niu2018VIPL} have nine scenarios, three RGB cameras, different illumination conditions, and different levels of movement. It is worth mentioning that the BVP signal and video of the dataset do not match in the time dimension. We used the output of HR head as the evaluation result. Besides, we normalize STMap to 30 fps by cubic spline interpolation to solve the problem of the unstable frame rate of video \cite{song2020heart,lu2021dual,yu2021physformer}.

% \vspace{0.1in} 
\noindent \textbf{V4V}~\cite{revanur2021V4V} is designed to collect data with the drastic changes of physiological indicators by simulating ten tasks such as a funny joke, 911 emergency call, and odor experience. It is worth mentioning that There are only heart rate labels in the dataset and no BVP signal labels. We used the output of HR head as the evaluation result.

% \vspace{0.1in} 
\noindent \textbf{BUAA}~\cite{xi2020BUAA} is proposed to evaluate the performance of the algorithm against various illumination. We only use data with illumination greater than or equal to 10 lux because underexposed images require special algorithms that are not considered in this article. We used the output of BVP head as the evaluation for BUAA, PURE, and UBFC-rPPG datasets.

% \vspace{0.1in} 
\noindent \textbf{PURE}~\cite{PURE2014} contains 60 RGB videos from 10 subjects with six different activities, specifically, sitting still, talking, and four rotating and moving head variations. The BVP signals are down-sampled from 60 to 30 fps with cubic spline interpolation to align with the videos.

% \vspace{0.1in} 
\noindent \textbf{UBFC-rPPG}~\cite{UBFC2017} containing 42 face videos with sunlight and indoor illumination, the ground-truth BVP signals, and HR values were collected by CMS50E.

\subsection{Implementation Details}
\label{sec:implement}
\begin{figure}[!t]
  \centering
   \includegraphics[width=0.95\linewidth]{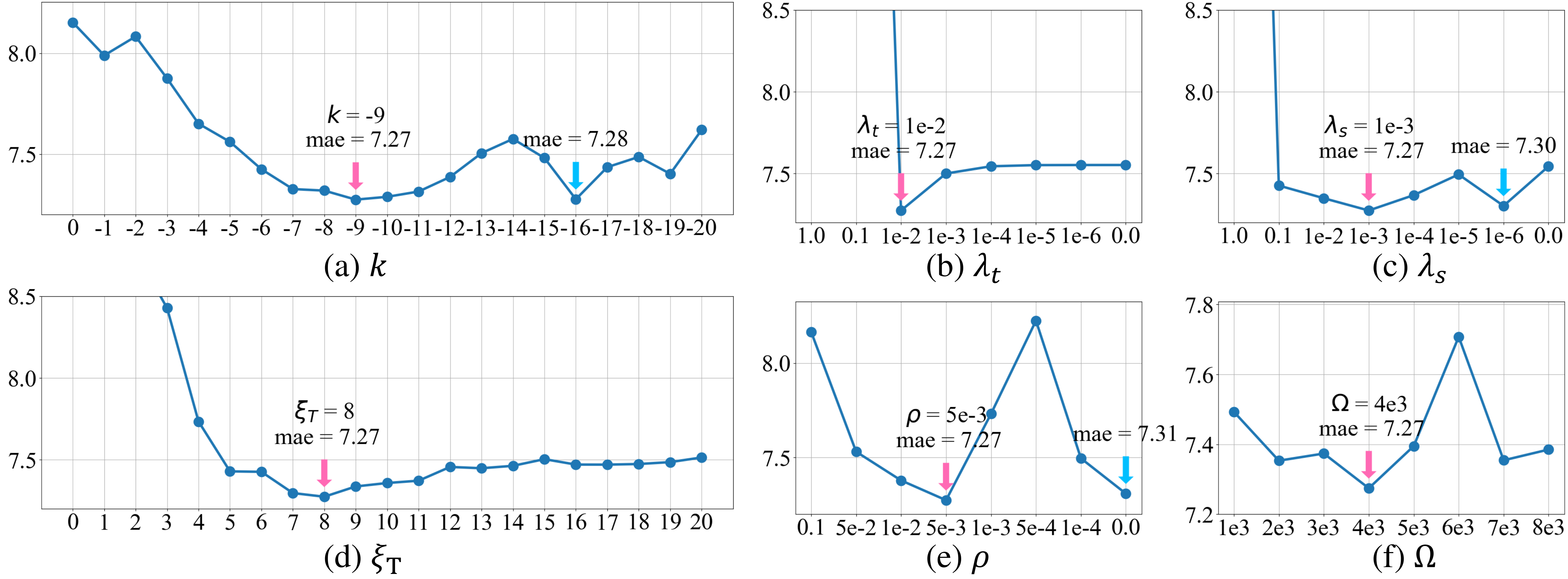}
   \caption{\textbf{Ablation experiments of hyper-parameters} on down-sampled VIPL dataset$\empty^{\ref{fn:vipl_20_interval}}$. Adheres to the default parameter configuration detailed in Sec.\ref{sec:implement}, only one hyper-parameter is varied during each set of experiments. The \textcolor{magenta}{pink} arrows point to the lowest MAE result and the \textcolor{cyan}{blue} arrows mark the second if indiscernible.}
   \label{fig:abl_hp}
\end{figure}

We implement our proposed priors and Bi-TTA using the PyTorch framework. The STMap sample is denoted as $\boldsymbol{x} \in \mathbb{R}^{256\times25\times3}$, where 256 is the temporal length, \textit{i.e.}, the number of video frames, of the sliding window, and 25 is the spatial dimension of the STMap. It is firstly resized to $\boldsymbol{x}' \in \mathbb{R}^{256\times64\times3}$ for network input. The network utilizes ResNet-18 \cite{he2016deep} as the base architecture, followed by an additional fully connected layer for HR estimation. During the TCL calculation, the  temporal perturbation $\delta_{\text{T}}$ is randomly sampled from a uniform distribution $\mathbb{U}_{(0,59]}$. Furthermore, we use stochastic gradient descent (SGD) with a momentum of 0.9 as the base optimizer. The bidirectional adaptation mechanism is implemented on top of the SGD.
% , based on which the bidirectional adaptation is implemented. 
During the adaptation process, we set the learning rate to 0.0001 and the batch size is 1. The hyper-parameters are configured as follows: $\lambda_{\text{s}}=0.001$, $\xi_{\text{T}}=8.0$, $\lambda_{\text{t}}=0.01$, $\rho=0.005$, $k=-9.0$, and $\Omega=4000$. Please refer to Fig.~\ref{fig:abl_hp} for detailed ablations. 

\footnotetext[5]{Note that the performance of down-sampled VIPL dataset might be inferior than the full VIPL dataset, due to the larger interval between neighboring samples.\label{fn:vipl_20_interval}}

\begin{figure}[!b]
\centering
\subcaptionbox{Comparison between priors and GT. \label{fig:abl_prior_gt}}{
\includegraphics[width = 0.41\linewidth]{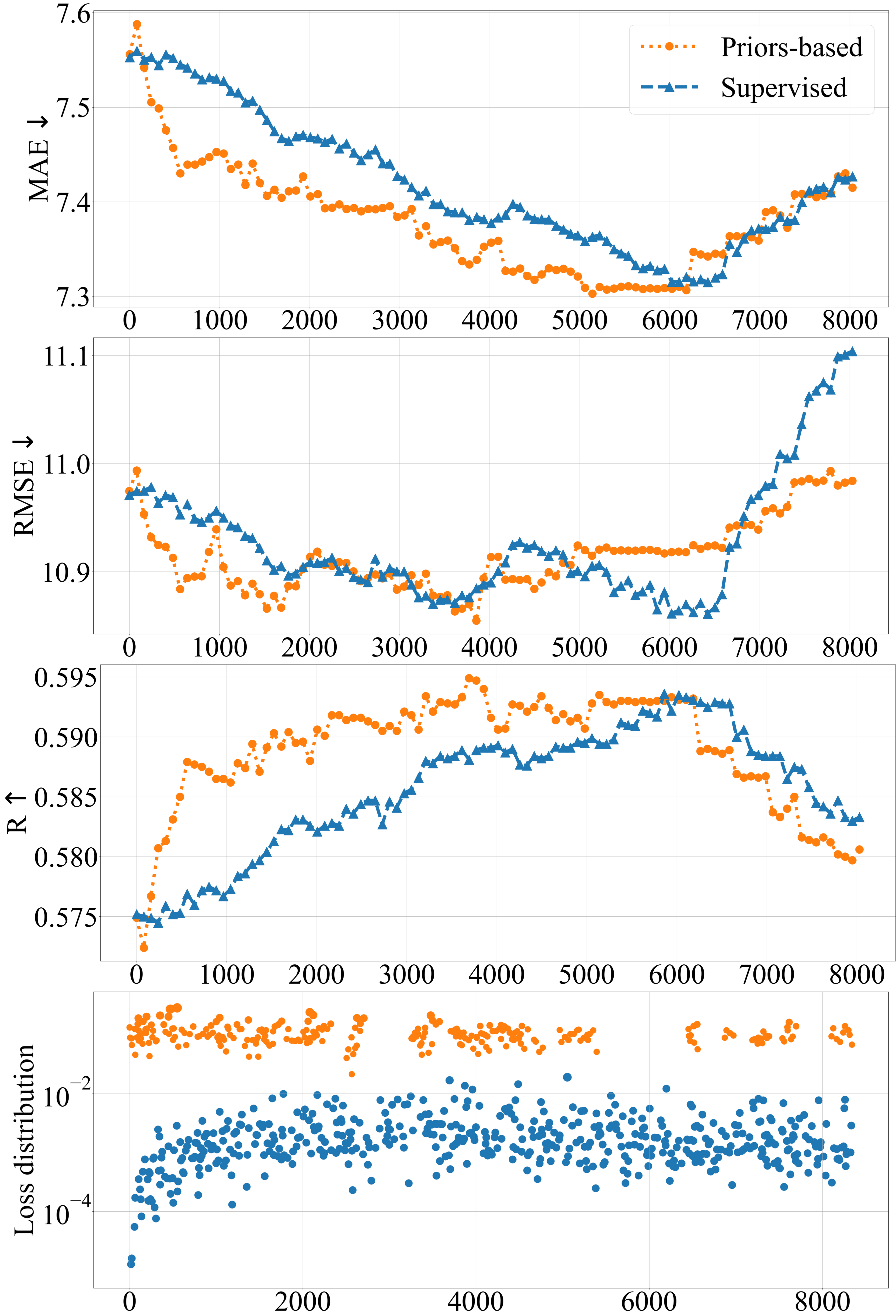}
}
\hfill
\subcaptionbox{Ablation study of PA, RS, and, priors.\label{fig:abl_all}}{
\includegraphics[width = 0.535\linewidth]{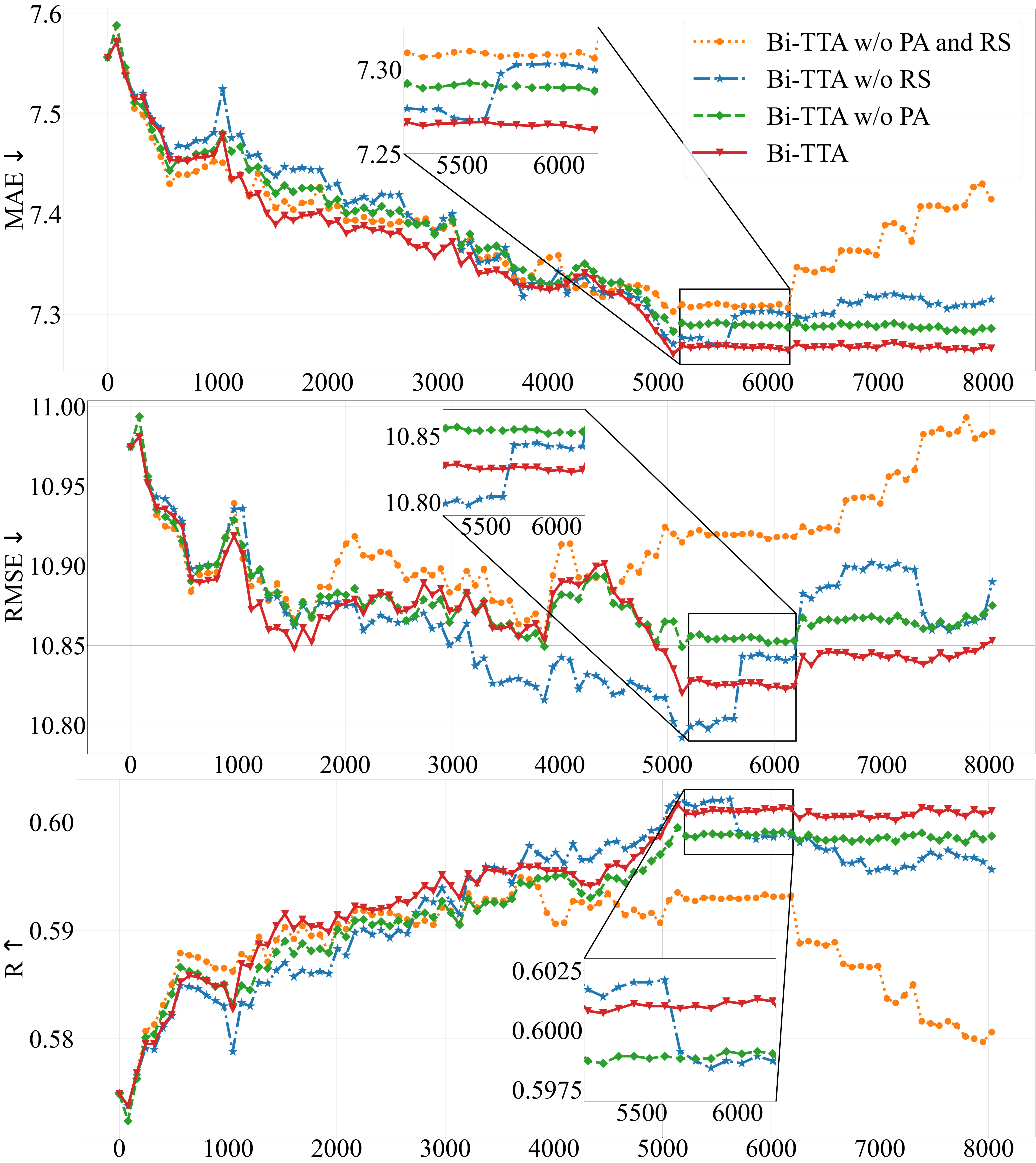}}
% \vspace{0.15cm}%final
\caption{\textbf{(a) Comparison between task-specific priors-based TTA and fully supervised adaptation.} The evaluation metrics are MSE, RMSE, and r. Alongside the visualization of the loss distribution for both priors-based and supervised adaptation method.
Loss values smaller than $10^{-6}$ are ignored. Note that for better comparison, the learning rate of supervised adaptation is set to 0.0005, while for the priors-based TTA it is 0.0001.
\textbf{(b) Ablation study of the adaptation performance among the proposed PA and RS modules in our Bi-TTA framework.} This Ablation study demonstrates the effectiveness of the PA and RS, in achieving a more effective and stable adaptation process.}
\label{fig:abl}
\end{figure}

\subsection{Results}
\label{sec:results}
Following existing rPPG methods \cite{Chen_2018_ECCV,song2020heart,yu2021physformer,CVD2020,lu2021dual,Lu_2023_CVPR}, mean absolute error (MAE), root mean square error (RMSE), and Pearson’s correlation coefficient (r) are used to evaluate the HR estimation, which inherently reflects the ability to extract physiological periodic signals. The results are organized and recorded in Tab.~\ref{tab:exp}. It is evident that TTA methods generally perform better than DG approaches. Among these TTA methods, our Bi-TTA demonstrates significantly superior performance, showcasing our effectiveness especially when both prospective and retrospective adaptations are synergistically employed. This highlights Bi-TTA's robustness in adapting to various unseen domains, underscoring its potential for real-world applications.

\subsection{Ablation Study}
\label{sec:ablation}
In this sub-section, we will mainly discuss about the effectiveness of the proposed priors and the bidirectional adaptation strategy.% , \textit{i.e.}, prospective and retrospective %  For the ablation study upon hyper-parameters, please see Fig.~\ref{fig:abl_hp}.

% \vspace{0.1in} 
\noindent \textbf{Task-specific priors.} As depicted in Fig.~\ref{fig:abl_prior_gt}, compared to ground-truth supervision, our task-specific priors provide effective adaptation supervision only on a significantly smaller number of samples, but induce considerably larger gradients. Thus, we can infer that the convergence rate under our self-supervised priors is markedly faster than ground-truth, despite a substantially lower learning rate. This phenomenon underscores the effectiveness of our priors.

As discussed in \cite{zhang2021understanding}, modern deep learning models are prone to memorize training data due to their over-parameterized design, leading to overfitting risks. Merely minimizing standard loss functions on training data often falls short in ensuring satisfactory generalization ability \cite{foret2021sharpnessaware}. Given that the rPPG model has undergone extensive pre-training with supervised learning, its parameters have likely been excessively optimized. In contrast, task knowledge-based priors are expected to offer fresh and correct guidance, steering the model towards improved generalization capability. % anticipated

% \vspace{0.5cm}
\noindent \textbf{Prospective and retrospective adaptation.} As depicted in Fig.~\ref{fig:abl_all}, the initial convergence rate and adaptation performance of the four comparisons are basically consistent across the first 4000 samples. However, in the latter half, the priors-based method demonstrates noticeable performance degradation. In contrast, other three methods, through utilizing prospective or retrospective adaptations, substantially mitigate this problem, thereby enhancing the adaptation ability.

Notably, between the 5000$^\text{th}$ and 6000$^\text{th}$ samples, while methods employing prospective adaptation exhibit rapid convergence and superior performance, they also display a significant degree of performance decay due to overfitting or catastrophic forgetting, which even intensifies as more samples are processed. On the other hand, the retrospective adaptation strategy ensures a more stable post-convergence maintenance in the target domain. The Bi-TTA, integrating both prospective and retrospective strategies, adeptly balances convergence speed, effectiveness, and stability.

\section{Conclusion}
\label{sec:Conclusion}
In this paper, we address the challenge of significant performance degradation when adapting rPPG models to unfamiliar target domains.
Considering privacy, we implement TTA for rPPG tasks, enabling the adaptation of pre-trained models to the target domain using only unannotated, instance-level target data. The models are fine-tuned during the inference of target samples.
In particular, we introduce Bi-TTA, a method that leverages spatial and temporal consistency for effective self-supervision, coupled with pioneering prospective and retrospective adaptation strategies. This methodology streamlines the adaptation process and outperforms existing techniques, demonstrating substantial promise for real-world applications in non-invasive physiological monitoring.

%% file: main.bbl
\begin{thebibliography}{10}
\providecommand{\url}[1]{\texttt{#1}}
\providecommand{\urlprefix}{URL }
\providecommand{\doi}[1]{https://doi.org/#1}

\bibitem{bahmani2022semantic}
Bahmani, S., Hahn, O., Zamfir, E., Araslanov, N., Cremers, D., Roth, S.: Semantic self-adaptation: Enhancing generalization with a single sample. arXiv preprint arXiv:2208.05788  (2022)

\bibitem{UBFC2017}
Bobbia, S., Macwan, R., Benezeth, Y., Mansouri, A., Dubois, J.: Unsupervised skin tissue segmentation for remote photoplethysmography. Pattern Recognition Letters  \textbf{124},  82--90 (2019)

\bibitem{brock2021high}
Brock, A., De, S., Smith, S.L., Simonyan, K.: High-performance large-scale image recognition without normalization. In: International Conference on Machine Learning. pp. 1059--1071. PMLR (2021)

\bibitem{Chen_2018_ECCV}
Chen, W., McDuff, D.: Deepphys: Video-based physiological measurement using convolutional attention networks. In: Proc. ECCV. pp. 349--365 (2018)

\bibitem{bvpnet}
Das, A., Lu, H., Han, H., Dantcheva, A., Shan, S., Chen, X.: Bvpnet: Video-to-bvp signal prediction for remote heart rate estimation. In: FG. pp. 01--08. IEEE (2021)

\bibitem{de2013robust_CHROM}
De~Haan, G., Jeanne, V.: Robust pulse rate from chrominance-based rppg. IEEE Trans. Biomed. Eng.  \textbf{60}(10),  2878--2886 (2013)

\bibitem{de2014improved}
De~Haan, G., Van~Leest, A.: Improved motion robustness of remote-ppg by using the blood volume pulse signature. Physiol. Meas.  \textbf{35}(9), ~1913 (2014)

\bibitem{d2019learning}
D'Innocente, A., Bucci, S., Caputo, B., Tommasi, T.: Learning to generalize one sample at a time with self-supervision. arXiv preprint arXiv:1910.03915  (2019)

\bibitem{Du_2023_CVPR}
Du, J., Liu, S.Q., Zhang, B., Yuen, P.C.: Dual-bridging with adversarial noise generation for domain adaptive rppg estimation. In: Proceedings of the IEEE/CVF Conference on Computer Vision and Pattern Recognition (CVPR). pp. 10355--10364 (June 2023)

\bibitem{du2023dual}
Du, J., Liu, S.Q., Zhang, B., Yuen, P.C.: Dual-bridging with adversarial noise generation for domain adaptive rppg estimation. In: Proceedings of the IEEE/CVF Conference on Computer Vision and Pattern Recognition. pp. 10355--10364 (2023)

\bibitem{d2020one}
D’Innocente, A., Borlino, F.C., Bucci, S., Caputo, B., Tommasi, T.: One-shot unsupervised cross-domain detection. In: Computer Vision--ECCV 2020: 16th European Conference, Glasgow, UK, August 23--28, 2020, Proceedings, Part XVI 16. pp. 732--748. Springer (2020)

\bibitem{pmlr-v70-finn17a}
Finn, C., Abbeel, P., Levine, S.: Model-agnostic meta-learning for fast adaptation of deep networks. In: Precup, D., Teh, Y.W. (eds.) Proceedings of the 34th International Conference on Machine Learning. Proceedings of Machine Learning Research, vol.~70, pp. 1126--1135. PMLR (06--11 Aug 2017), \url{https://proceedings.mlr.press/v70/finn17a.html}

\bibitem{foret2021sharpnessaware}
Foret, P., Kleiner, A., Mobahi, H., Neyshabur, B.: Sharpness-aware minimization for efficiently improving generalization. In: International Conference on Learning Representations (2021), \url{https://openreview.net/forum?id=6Tm1mposlrM}

\bibitem{gandelsman2022testtime}
Gandelsman, Y., Sun, Y., Chen, X., Efros, A.A.: Test-time training with masked autoencoders. In: Oh, A.H., Agarwal, A., Belgrave, D., Cho, K. (eds.) Advances in Neural Information Processing Systems (2022), \url{https://openreview.net/forum?id=SHMi1b7sjXk}

\bibitem{gao2023back}
Gao, J., Zhang, J., Liu, X., Darrell, T., Shelhamer, E., Wang, D.: Back to the source: Diffusion-driven adaptation to test-time corruption. In: Proceedings of the IEEE/CVF Conference on Computer Vision and Pattern Recognition. pp. 11786--11796 (2023)

\bibitem{hsu2017deep}
Gee-Sern~Hsu, ArulMurugan~Ambikapathi, M.S.C.: Deep learning with time-frequency representation for pulse estimation. In: Proc. IJCB. pp. 642--650 (2017)

\bibitem{gideon2021way}
Gideon, J., Stent, S.: The way to my heart is through contrastive learning: Remote photoplethysmography from unlabelled video. In: Proceedings of the IEEE/CVF international conference on computer vision. pp. 3995--4004 (2021)

\bibitem{he2016deep}
He, K., Zhang, X., Ren, S., Sun, J.: Deep residual learning for image recognition. In: Proc. IEEE CVPR. pp. 770--778 (2016)

\bibitem{jia2021scaling}
Jia, C., Yang, Y., Xia, Y., Chen, Y.T., Parekh, Z., Pham, H., Le, Q., Sung, Y.H., Li, Z., Duerig, T.: Scaling up visual and vision-language representation learning with noisy text supervision. In: International conference on machine learning. pp. 4904--4916. PMLR (2021)

\bibitem{kessler2017pain}
Kessler, V., Thiam, P., Amirian, M., Schwenker, F.: Pain recognition with camera photoplethysmography. In: IEEE IPTA. pp.~1--5 (2017)

\bibitem{khurana2021sita}
Khurana, A., Paul, S., Rai, P., Biswas, S., Aggarwal, G.: Sita: Single image test-time adaptation. arXiv preprint arXiv:2112.02355  (2021)

\bibitem{klingner2022continual}
Klingner, M., Ayache, M., Fingscheidt, T.: Continual batchnorm adaptation (cbna) for semantic segmentation. IEEE Transactions on Intelligent Transportation Systems  \textbf{23}(11),  20899--20911 (2022)

\bibitem{VREx}
Krueger, D., Caballero, E., Jacobsen, J.H., Zhang, A., Binas, J., Zhang, D., Le~Priol, R., Courville, A.: Out-of-distribution generalization via risk extrapolation (rex). In: ICML. pp. 5815--5826. PMLR (2021)

\bibitem{lee2020meta}
Lee, E., Chen, E., Lee, C.Y.: Meta-rppg: Remote heart rate estimation using a transductive meta-learner. In: Computer Vision--ECCV 2020: 16th European Conference, Glasgow, UK, August 23--28, 2020, Proceedings, Part XXVII 16. pp. 392--409. Springer (2020)

\bibitem{2011rPPGPCA}
Lewandowska, M., Rumi{\'n}ski, J., Kocejko, T., Nowak, J.: Measuring pulse rate with a webcam—a non-contact method for evaluating cardiac activity. In: FedCSIS. pp. 405--410. IEEE (2011)

\bibitem{li2018visualizing}
Li, H., Xu, Z., Taylor, G., Studer, C., Goldstein, T.: Visualizing the loss landscape of neural nets. Advances in neural information processing systems  \textbf{31} (2018)

\bibitem{liang2023comprehensive}
Liang, J., He, R., Tan, T.: A comprehensive survey on test-time adaptation under distribution shifts. arXiv preprint arXiv:2303.15361  (2023)

\bibitem{liang2020we}
Liang, J., Hu, D., Feng, J.: Do we really need to access the source data? source hypothesis transfer for unsupervised domain adaptation. In: International Conference on Machine Learning (ICML). pp. 6028--6039 (2020)

\bibitem{liang2021source}
Liang, J., Hu, D., Wang, Y., He, R., Feng, J.: Source data-absent unsupervised domain adaptation through hypothesis transfer and labeling transfer. IEEE Transactions on Pattern Analysis and Machine Intelligence (TPAMI)  (2021), in Press

\bibitem{liu2021metaphys}
Liu, X., Jiang, Z., Fromm, J., Xu, X., Patel, S., McDuff, D.: Metaphys: few-shot adaptation for non-contact physiological measurement. In: Proceedings of the conference on health, inference, and learning. pp. 154--163 (2021)

\bibitem{liu2023rppg}
Liu, X., Zhang, Y., Yu, Z., Lu, H., Yue, H., Yang, J.: rppg-mae: Self-supervised pre-training with masked autoencoders for remote physiological measurement. arXiv preprint arXiv:2306.02301  (2023)

\bibitem{liu2021ttt++}
Liu, Y., Kothari, P., van Delft, B.G., Bellot-Gurlet, B., Mordan, T., Alahi, A.: Ttt++: When does self-supervised test-time training fail or thrive? In: Thirty-Fifth Conference on Neural Information Processing Systems (2021)

\bibitem{lu2021dual}
Lu, H., Han, H., Zhou, S.K.: Dual-gan: Joint bvp and noise modeling for remote physiological measurement. In: Proc. IEEE CVPR. pp. 12404--12413 (2021)

\bibitem{Lu_2023_CVPR}
Lu, H., Yu, Z., Niu, X., Chen, Y.C.: Neuron structure modeling for generalizable remote physiological measurement. In: Proceedings of the IEEE/CVF Conference on Computer Vision and Pattern Recognition (CVPR). pp. 18589--18599 (June 2023)

\bibitem{mcduff2021camera}
McDuff, D.: Camera measurement of physiological vital signs. CSUR  (2021)

\bibitem{mcduff2022applications}
McDuff, D.: Applications of camera-based physiological measurement beyond healthcare. In: CVSM, pp. 165--177. Elsevier (2022)

\bibitem{2014ICA}
McDuff, D., Gontarek, S., Picard, R.W.: Improvements in remote cardiopulmonary measurement using a five band digital camera. IEEE Trans. Biomed. Eng.  \textbf{61}(10),  2593--2601 (2014)

\bibitem{mi2022make}
Mi, P., Shen, L., Ren, T., Zhou, Y., Sun, X., Ji, R., Tao, D.: Make sharpness-aware minimization stronger: A sparsified perturbation approach (2022)

\bibitem{niu2022efficient}
Niu, S., Wu, J., Zhang, Y., Chen, Y., Zheng, S., Zhao, P., Tan, M.: Efficient test-time model adaptation without forgetting. In: The Internetional Conference on Machine Learning (2022)

\bibitem{niu2023towards}
Niu, S., Wu, J., Zhang, Y., Wen, Z., Chen, Y., Zhao, P., Tan, M.: Towards stable test-time adaptation in dynamic wild world. In: Internetional Conference on Learning Representations (2023)

\bibitem{synrhythm}
Niu, X., Han, H., Shan, S., Chen, X.: Synrhythm: Learning a deep heart rate estimator from general to specific. In: Proc. IEEE ICPR. pp. 3580--3585 (2018)

\bibitem{niu2018VIPL}
Niu, X., Han, H., Shan, S., Chen, X.: {VIPL-HR}: A multi-modal database for pulse estimation from less-constrained face video. In: Proc. ACCV. pp. 562--576 (2018)

\bibitem{niu2019robust}
Niu, X., Shan, S., Han, H., Chen, X.: Rhythmnet: End-to-end heart rate estimation from face via spatial-temporal representation. IEEE Trans. on Image Process.  \textbf{29},  2409--2423 (2020)

\bibitem{CVD2020}
Niu, X., Yu, Z., Han, H., Li, X., Shan, S., Zhao, G.: Video-based remote physiological measurement via cross-verified feature disentangling. In: Proc. ECCV (2020)

\bibitem{pham2021meta}
Pham, H., Dai, Z., Xie, Q., Le, Q.V.: Meta pseudo labels. In: Proceedings of the IEEE/CVF conference on computer vision and pattern recognition. pp. 11557--11568 (2021)

\bibitem{poh2010non}
Poh, M.Z., McDuff, D.J., Picard, R.W.: Non-contact, automated cardiac pulse measurements using video imaging and blind source separation. Opt. Express  \textbf{18}(10),  10762--10774 (2010)

\bibitem{Qiu2019EVM}
Qiu, Y., Liu, Y., Arteaga-Falconi, J., Dong, H., El~Saddik, A.: Evm-cnn: Real-time contactless heart rate estimation from facial video. IEEE Trans. Multimedia  (2019)

\bibitem{reiss2019deep}
Reiss, A., Indlekofer, I., Schmidt, P., Van~Laerhoven, K.: Deep ppg: large-scale heart rate estimation with convolutional neural networks. Sensors  \textbf{19}(14), ~3079 (2019)

\bibitem{revanur2021V4V}
Revanur, A., Li, Z., Ciftci, U.A., Yin, L., Jeni, L.A.: The first vision for vitals (v4v) challenge for non-contact video-based physiological estimation. In: Proc. CVPR workshop. pp. 2760--2767 (2021)

\bibitem{2020PulseGAN}
Song, R., Chen, H., Cheng, J., Li, C., Liu, Y., Chen, X.: Pulsegan: Learning to generate realistic pulse waveforms in remote photoplethysmography. IEEE J-BHI pp.~1--1 (2021)

\bibitem{song2020heart}
Song, R., Zhang, S., Li, C., Zhang, Y., Cheng, J., Chen, X.: Heart rate estimation from facial videos using a spatiotemporal representation with convolutional neural networks. IEEE Trans. Instrum. Meas.  (2020)

\bibitem{vspetlik2018visual}
{\v{S}}petl{\'\i}k, R., Franc, V., Matas, J.: Visual heart rate estimation with convolutional neural network. In: Proc. BMVC. pp.~3--6 (2018)

\bibitem{PURE2014}
Stricker, R., M{\"u}ller, S., Gross, H.M.: Non-contact video-based pulse rate measurement on a mobile service robot. In: Proc. IEEE ISRHIC. pp. 1056--1062 (2014)

\bibitem{coral}
Sun, B., Saenko, K.: Deep coral: Correlation alignment for deep domain adaptation. In: ECCV. pp. 443--450. Springer (2016)

\bibitem{sun2016lstm}
Sun, B., Wei, Q., Li, L., Xu, Q., He, J., Yu, L.: Lstm for dynamic emotion and group emotion recognition in the wild. In: Proc. ACM ICMI. pp. 451--457 (2016)

\bibitem{sun2023resolve}
Sun, W., Zhang, X., Lu, H., Chen, Y., Ge, Y., Huang, X., Yuan, J., Chen, Y.: Resolve domain conflicts for generalizable remote physiological measurement. In: Proceedings of the 31st ACM International Conference on Multimedia. pp. 8214--8224 (2023)

\bibitem{sun2020test}
Sun, Y., Wang, X., Liu, Z., Miller, J., Efros, A., Hardt, M.: Test-time training with self-supervision for generalization under distribution shifts. In: International conference on machine learning. pp. 9229--9248. PMLR (2020)

\bibitem{sun2022contrast}
Sun, Z., Li, X.: Contrast-phys: Unsupervised video-based remote physiological measurement via spatiotemporal contrast. In: European Conference on Computer Vision. pp. 492--510. Springer (2022)

\bibitem{NC2022PAMI}
Tian, C.X., Li, H., Xie, X., Liu, Y., Wang, S.: Neuron coverage-guided domain generalization. IEEE Trans. on Pattern Anal. Mach. Intell.  (2022)

\bibitem{verkruysse2008remote_GREEN}
Verkruysse, W., Svaasand, L.O., Nelson, J.S.: Remote plethysmographic imaging using ambient light. Opt. Express  \textbf{16}(26),  21434--21445 (2008)

\bibitem{wang2021tent}
Wang, D., Shelhamer, E., Liu, S., Olshausen, B., Darrell, T.: Tent: Fully test-time adaptation by entropy minimization. In: International Conference on Learning Representations (2021), \url{https://openreview.net/forum?id=uXl3bZLkr3c}

\bibitem{wang2022self}
Wang, H., Ahn, E., Kim, J.: Self-supervised representation learning framework for remote physiological measurement using spatiotemporal augmentation loss. In: Proceedings of the AAAI Conference on Artificial Intelligence. vol.~36, pp. 2431--2439 (2022)

\bibitem{wang2022generalizing}
Wang, J., Lan, C., Liu, C., Ouyang, Y., Qin, T., Lu, W., Chen, Y., Zeng, W., Yu, P.: Generalizing to unseen domains: A survey on domain generalization. IEEE Transactions on Knowledge and Data Engineering  (2022)

\bibitem{wang2017algorithmic_POS}
Wang, W., den Brinker, A.C., Stuijk, S., de~Haan, G.: Algorithmic principles of remote ppg. IEEE Trans. Biomed. Eng.  \textbf{64}(7),  1479--1491 (2017)

\bibitem{wang2015exploiting}
Wang, W., Stuijk, S., De~Haan, G.: Exploiting spatial redundancy of image sensor for motion robust rppg. IEEE Trans. Biomed. Eng.  \textbf{62}(2),  415--425 (2015)

\bibitem{wang2019vision}
Wang, Z.K., Kao, Y., Hsu, C.T.: Vision-based heart rate estimation via a two-stream cnn. In: Proc. IEEE ICIP. pp. 3327--3331 (2019)

\bibitem{wu2020adversarial}
Wu, D., Xia, S.T., Wang, Y.: Adversarial weight perturbation helps robust generalization. Advances in Neural Information Processing Systems  \textbf{33},  2958--2969 (2020)

\bibitem{xi2020BUAA}
Xi, L., Chen, W., Zhao, C., Wu, X., Wang, J.: Image enhancement for remote photoplethysmography in a low-light environment. In: FG. pp.~1--7. IEEE (2020)

\bibitem{yu2019remote}
Yu, Z., Peng, W., Li, X., Hong, X., Zhao, G.: Remote heart rate measurement from highly compressed facial videos: an end-to-end deep learning solution with video enhancement. In: Proc. IEEE ICCV. pp. 151--160 (2019)

\bibitem{yu2021physformer}
Yu, Z., Shen, Y., Shi, J., Zhao, H., Torr, P., Zhao, G.: Physformer: Facial video-based physiological measurement with temporal difference transformer. IEEE CVPR  (2022)

\bibitem{zhang2021understanding}
Zhang, C., Bengio, S., Hardt, M., Recht, B., Vinyals, O.: Understanding deep learning (still) requires rethinking generalization. Communications of the ACM  \textbf{64}(3),  107--115 (2021)

\bibitem{zhang2023adaptive}
Zhang, X., Chen, Y.C.: Adaptive domain generalization via online disagreement minimization. IEEE Transactions on Image Processing  (2023)

\bibitem{zhao2022test}
Zhao, X., Liu, C., Sicilia, A., Hwang, S.J., Fu, Y.: Test-time fourier style calibration for domain generalization. arXiv preprint arXiv:2205.06427  (2022)

\bibitem{zou2022learning}
Zou, Y., Zhang, Z., Li, C.L., Zhang, H., Pfister, T., Huang, J.B.: Learning instance-specific adaptation for cross-domain segmentation. In: European Conference on Computer Vision. pp. 459--476. Springer (2022)

\end{thebibliography}
